%% file: arxiv.tex
\documentclass{article}
\usepackage[letterpaper, left=1in, right=1in, top=1in, bottom=1in]{geometry}
\usepackage{natbib}
\bibpunct{(}{)}{;}{a}{,}{,}
\usepackage[colorlinks,citecolor=blue!70!black,linkcolor=blue!70!black,
urlcolor=blue!70!black,breaklinks=true]{hyperref}
\usepackage{parskip}
\usepackage[usenames, dvipsnames]{xcolor}
\usepackage{microtype}
\usepackage{booktabs}
\usepackage{amsmath}
\usepackage{dsfont}
\usepackage{amssymb}
\usepackage{xspace}
\usepackage{url}
\usepackage{algorithm}
\usepackage[noend]{algorithmic}
\usepackage{amsthm}
\usepackage{listings}
\usepackage{bm}
\usepackage{bbm}
\usepackage{enumitem}
\usepackage{nicefrac}
\usepackage{mathrsfs}
\usepackage{inconsolata}
\usepackage[subrefformat=parens,labelformat=parens]{subfig}
\usepackage{wrapfig}
\usepackage{graphicx}
\usepackage{bigints}
\usepackage{transparent}
\usepackage{comment}
\usepackage[nameinlink,capitalize]{cleveref}
\makeatletter
\AddToHook{cmd/appendix/before}{
\def\cref@section@alias{appendix}
\def\cref@subsection@alias{appendix}
\def\cref@subsubsection@alias{appendix}
}
\makeatother
\usepackage{thmtools}
\usepackage{thm-restate}
\usepackage{nicematrix}
\usepackage{arydshln}
\usepackage{fancyhdr}
\usepackage{mathtools}
\usepackage[scaled=.88]{helvet}
\usepackage{breakcites}
\usepackage{multirow}
\usepackage[most]{tcolorbox}

\input{macros}

\title{Test-Time Scaling for Diffusion Language Models with Frequency-Aware Remasking}
\date{}

\author{
Bowen Zuo$^{1}$\thanks{Equal contribution.}\\
{\small\texttt{bowen.zuo@email.ucr.edu}}
\and
Yue Yu$^{2}$\footnotemark[1]\\
{\small\texttt{yyu3@iu.edu}}
\and
Dongruo Zhou$^{2}$\\
{\small\texttt{dz13@iu.edu}}
\and
Yinglun Zhu$^{1}$\\
{\small\texttt{yzhu@ucr.edu}}
\and
~\\
{\normalsize $^{1}$University of California, Riverside \quad
$^{2}$Indiana University}
}

\begin{document}

\maketitle

\begin{abstract}
  \input{sections/abstract.tex}
\end{abstract}

\input{sections/intro.tex}

\input{sections/related.tex}

\input{sections/setting.tex}
\input{sections/methods.tex}

\input{sections/experiments.tex}

\input{sections/discussion.tex}

\newpage
\bibliography{refs}

\newpage
\appendix
\input{sections/appendix.tex}

\end{document}

%% file: macros.tex
\DeclareMathOperator*{\argmax}{arg\,max}

\renewcommand{\epsilon}{\varepsilon}

\newtheoremstyle{spaced}
  {6pt}   %
  {0pt}   %
  {\itshape} %
  {}       %
  {\bfseries} %
  {.}      %
  {0.5em}  %
  {}
\theoremstyle{spaced}

\newcommand{\algcommentlight}[1]{\textcolor{blue!70!black}{\transparent{0.5}\small{\texttt{\textbf{//\hspace{2pt}#1}}}}}

\DeclarePairedDelimiter{\abs}{\lvert}{\rvert} %

\DeclarePairedDelimiterX{\infdiv}[2]{(}{)}{%
  #1\;\delimsize\|\;#2%
}

\def\ddefloop#1{\ifx\ddefloop#1\else\ddef{#1}\expandafter\ddefloop\fi}
\def\ddef#1{\expandafter\def\csname bb#1\endcsname{\ensuremath{\mathbb{#1}}}}
\ddefloop ABCDEFGHIJKLMNOPQRSTUVWXYZ\ddefloop
\def\ddefloop#1{\ifx\ddefloop#1\else\ddef{#1}\expandafter\ddefloop\fi}
\def\ddef#1{\expandafter\def\csname b#1\endcsname{\ensuremath{\mathbf{#1}}}}
\ddefloop ABCDEFGHIJKLMNOPQRSTUVWXYZ\ddefloop
\def\ddef#1{\expandafter\def\csname sf#1\endcsname{\ensuremath{\mathsf{#1}}}}
\ddefloop ABCDEFGHIJKLMNOPQRSTUVWXYZ\ddefloop
\def\ddef#1{\expandafter\def\csname c#1\endcsname{\ensuremath{\mathcal{#1}}}}
\ddefloop ABCDEFGHIJKLMNOPQRSTUVWXYZ\ddefloop
\def\ddef#1{\expandafter\def\csname h#1\endcsname{\ensuremath{\widehat{#1}}}}
\ddefloop ABCDEFGHIJKLMNOPQRSTUVWXYZ\ddefloop
\def\ddef#1{\expandafter\def\csname hc#1\endcsname{\ensuremath{\widehat{\mathcal{#1}}}}}
\ddefloop ABCDEFGHIJKLMNOPQRSTUVWXYZ\ddefloop
\def\ddef#1{\expandafter\def\csname t#1\endcsname{\ensuremath{\widetilde{#1}}}}
\ddefloop ABCDEFGHIJKLMNOPQRSTUVWXYZ\ddefloop
\def\ddef#1{\expandafter\def\csname tc#1\endcsname{\ensuremath{\widetilde{\mathcal{#1}}}}}
\ddefloop ABCDEFGHIJKLMNOPQRSTUVWXYZ\ddefloop
\def\ddefloop#1{\ifx\ddefloop#1\else\ddef{#1}\expandafter\ddefloop\fi}
\def\ddef#1{\expandafter\def\csname scr#1\endcsname{\ensuremath{\mathscr{#1}}}}
\ddefloop ABCDEFGHIJKLMNOPQRSTUVWXYZ\ddefloop

\let\oldparagraph\paragraph
\renewcommand{\paragraph}[1]{\oldparagraph{#1}}

\renewcommand{\epsilon}{\varepsilon}

\renewcommand{\bigm}[1]{%
  \ifcsname fenced@\string#1\endcsname
    \expandafter\@firstoftwo
  \else
    \expandafter\@secondoftwo
  \fi
  {\expandafter\amsmath@bigm\csname fenced@\string#1\endcsname}%
  {\amsmath@bigm#1}%
}

\newcommand{\DeclareFence}[2]{\@namedef{fenced@\string#1}{#2}}
\makeatother

\makeatletter
\let\save@mathaccent\mathaccent
\newcommand*\if@single[3]{%
  \setbox0\hbox{${\mathaccent"0362{#1}}^H$}%
  \setbox2\hbox{${\mathaccent"0362{\kern0pt#1}}^H$}%
  \ifdim\ht0=\ht2 #3\else #2\fi
  }
\newcommand*\rel@kern[1]{\kern#1\dimexpr\macc@kerna}
\newcommand*\widebar[1]{\@ifnextchar^{{\wide@bar{#1}{0}}}{\wide@bar{#1}{1}}}
\newcommand*\wide@bar[2]{\if@single{#1}{\wide@bar@{#1}{#2}{1}}{\wide@bar@{#1}{#2}{2}}}
\newcommand*\wide@bar@[3]{%
  \begingroup
  \def\mathaccent##1##2{%
    \let\mathaccent\save@mathaccent
    \if#32 \let\macc@nucleus\first@char \fi
    \setbox\z@\hbox{$\macc@style{\macc@nucleus}_{}$}%
    \setbox\tw@\hbox{$\macc@style{\macc@nucleus}{}_{}$}%
    \dimen@\wd\tw@
    \advance\dimen@-\wd\z@
    \divide\dimen@ 3
    \@tempdima\wd\tw@
    \advance\@tempdima-\scriptspace
    \divide\@tempdima 10
    \advance\dimen@-\@tempdima
    \ifdim\dimen@>\z@ \dimen@0pt\fi
    \rel@kern{0.6}\kern-\dimen@
    \if#31
      \overline{\rel@kern{-0.6}\kern\dimen@\macc@nucleus\rel@kern{0.4}\kern\dimen@}%
      \advance\dimen@0.4\dimexpr\macc@kerna
      \let\final@kern#2%
      \ifdim\dimen@<\z@ \let\final@kern1\fi
      \if\final@kern1 \kern-\dimen@\fi
    \else
      \overline{\rel@kern{-0.6}\kern\dimen@#1}%
    \fi
  }%
  \macc@depth\@ne
  \let\math@bgroup\@empty \let\math@egroup\macc@set@skewchar
  \mathsurround\z@ \frozen@everymath{\mathgroup\macc@group\relax}%
  \macc@set@skewchar\relax
  \let\mathaccentV\macc@nested@a
  \if#31
    \macc@nested@a\relax111{#1}%
  \else
    \def\gobble@till@marker##1\endmarker{}%
    \futurelet\first@char\gobble@till@marker#1\endmarker
    \ifcat\noexpand\first@char A\else
      \def\first@char{}%
    \fi
    \macc@nested@a\relax111{\first@char}%
  \fi
  \endgroup
}
\makeatother

%% file: sections/abstract.tex
Diffusion language models (DLMs) generate text through iterative denoising, where the choice of which tokens to commit and which to remask drives output quality. Existing certainty-aware strategies rely on instantaneous confidence or entropy to guide this choice, but these local signals can be misleading on hard reasoning problems. We therefore formulate DLM test-time scaling as an adaptive allocation over both compute and decoding strategy. Our two-stage method uses a fast base sampler to solve easy questions, then applies frequency-aware remasking to the remaining hard questions. Frequency-aware remasking tracks how often each token's prediction changes across denoising steps and commits the frequently changing tokens first, so that samples diverge at the undecided positions rather than collapsing onto the model's preferred reasoning path. Experiments with LLaDA-8B-Instruct and Dream-7B-Instruct show consistent gains over a strong adaptive allocation baseline: up to 3.73\% (LLaDA) and 7.33\% (Dream) absolute accuracy improvements on MATH-500, and up to 11.39\% (LLaDA) and 10.98\% (Dream) absolute coverage improvements on HumanEval. On questions that remain unsolved by the base certainty-aware strategy even after 100 samples, frequency-aware remasking recovers solutions for 3.75\% to 18.75\% of examples.

%% file: sections/intro.tex
\section{Introduction}
\label{sec:intro}

Diffusion language models (DLMs) have emerged as a promising alternative to autoregressive language models for text generation. Rather than generating tokens strictly from left to right, DLMs produce sequences through iterative denoising, where masked tokens are gradually unmasked into a complete response. This direction builds on discrete and masked diffusion formulations for language \citep{austin2021structured,hoogeboom2021argmax,li2022diffusionlm,sahoo2024simple}, and recent large-scale DLMs such as LLaDA and Dream perform competitively with similarly sized autoregressive models \citep{nie2025large,ye2025dream}.\looseness=-1

A key design choice in DLM inference is the token update or remasking strategy. Existing methods typically use the model's instantaneous predictive distribution to decide which tokens should be fixed and which should remain available for further processing \citep{nie2025large,li2026diffusion}. For example, LLaDA uses confidence-driven remasking, where high-confidence tokens are progressively unmasked while low-confidence tokens remain masked longer \citep{nie2025large}. DREAM-style implementations commonly use entropy-based remasking, where high-entropy positions are treated as uncertain and kept active for additional denoising \citep{ben-hamu2025accelerated,gong2026diffucoder}. Recent fast inference frameworks further accelerate DLM decoding through confidence-thresholding or confidence-aware parallel decoding, committing tokens once their confidence exceeds a predefined threshold \citep{wu2026fastdllm,ma2025dinfer,mohamed2025fast}. These approaches improve inference efficiency, but they share a common principle: local certainty under the current predictive distribution is used as a proxy for whether a token has been correctly resolved.

The choice of decoding strategy has a substantial effect on DLM performance. For \texttt{Dream-7B-Instruct} under a fixed 256-step budget, switching the decoding heuristic from low-confidence to entropy changes MATH-500 accuracy from $21.80\%$ to $32.20\%$ \citep{li2026breakingfactorizationbarrierdiffusion}, and reported \texttt{LLaDA-8B-Instruct} baselines on MATH-500 range from roughly $26\%$ to $42\%$ depending on generation length, denoising steps, and decoding strategy \citep{nie2025large,bai2026prismefficienttesttimescaling,feng2026dvotingfastvotingdllms,zhao2025d}. DLM performance should therefore not be characterized by a single fixed decoding run. More importantly, these observations point to a structural property of DLM decoding: different strategies can reach similar aggregate accuracy while solving different subsets of questions.
\looseness=-1

This complementary view motivates an adaptive view of DLM decoding. Rather than applying a fixed sampler uniformly to all inputs, the decoder should identify which questions a fast strategy solves and which need a different decoding behavior. We therefore study DLM test-time scaling as an adaptive allocation problem over both \textbf{number of responses} and \textbf{decoding strategy}. The remaining question is what that second decoding behavior should be.

Within this framing, we identify \emph{false confidence} as a limitation of certainty-aware decoding: tokens can appear locally confident while still belonging to an incorrect reasoning trajectory. On hard reasoning problems, these confident commitments quietly steer every sample toward the model's preferred reasoning path, so repeated samples from the same certainty-aware decoder converge to the same wrong answer. Local confidence and entropy alone cannot diagnose this failure, because they are snapshots of a single denoising step. To address this limitation, we propose frequency-aware remasking, which augments certainty-aware remasking with a trajectory-level signal based on how often token predictions change across denoising steps and reverses the usual commitment order: frequently changing tokens are committed first, so that different samples make different choices at the positions where the model is still undecided and go on to explore different solution paths. We incorporate this signal into a two-stage adaptive method: a fast base sampler first resolves easy questions, and frequency-aware remasking is then applied only to the remaining unresolved questions. We evaluate our approach on mathematical reasoning and code generation benchmarks, examining both standard test-time scaling behavior and the recovery of challenging questions left unresolved by certainty-based decoding. The results show that adaptive allocation and frequency-aware remasking improve the efficiency--accuracy tradeoff by combining the speed of certainty-based decoding with the robustness of frequency-aware remasking.
\looseness=-1

Our contributions are summarized as follows:
\begin{itemize}
    \item We formulate DLM test-time scaling as an adaptive allocation problem over the \textbf{number of responses} and \textbf{decoding strategies}, and propose a two-stage algorithm that switches decoding strategy for the questions the fast base sampler fails to solve.

    \item We identify \emph{false confidence} as a motivation for trajectory-level decoding signals, and introduce frequency-aware remasking to complement instantaneous confidence- and entropy-based token update rules.

    \item We show that frequency-aware remasking recovers solutions for up to $18.75\%$ of challenging reasoning questions, while also yielding up to $7.33\%$ absolute accuracy gains on MATH-500 and $11.39\%$ coverage gains on HumanEval.
\end{itemize}

%% file: sections/related.tex
\section{Related Work}
\label{sec:related}

\subsection{Test-Time Scaling for Language Models}
Test-time compute scaling has proven effective for autoregressive models via best-of-$N$ sampling, beam search, chain-of-thought, tree search, and adaptive allocation \citep{brown2024largelanguagemonkeysscaling, snell2025scaling, wei2023chainofthoughtpromptingelicitsreasoning, yao2023tree, levi2025a, liu2025can, yu2026on}. Adapting these ideas to diffusion language models is less straightforward, as the natural scaling axis—denoising steps—is often already saturated. \citet{wang2025remasking} establish a formal connection between remasking and inference-time scaling, showing that increasing sampling steps monotonically improves quality. Subsequent work scales along complementary axes: Prism \citep{bai2026prismefficienttesttimescaling} combines hierarchical search with self-verification; \citet{bilal2026s} searches over denoising trajectories; RFG \citep{chen2026rfg} provides reward-free guidance via log-likelihood ratios.

These search-based methods scale compute along the width axis. For AR models, adaptive allocation---deciding where to spend compute rather than how many samples to draw---is a complementary and important scaling dimension \citep{wang-etal-2025-make, zuo2026strategic,damani2025learning, fu2025efficiently, zuo-etal-2026-adaptive}. In the DLM setting, this dimension remains underexplored. \citet{liang2026cd4lmconsistencydistillationadaptive} identify rigid unmasking schedules as a key limitation, and propose adaptive NFE allocation through a redesigned training objective. \citet{cholakov2026optimizing} address the same problem at the scheduling level, learning via RL which token positions to denoise at each step. Our work targets this allocation problem from a complementary angle: rather than changing the training objective or learning a token-level denoising schedule, we adaptively allocate questions across decoding strategies at inference time.
\looseness=-1

\subsection{Remasking Strategy and Efficient Sampling in Diffusion Language Models}

A central design choice in DLM inference is the token update rule: which tokens should be fixed, and which should remain available for refinement. Standard masked diffusion decoders often treat unmasking as effectively irreversible: once a token is accepted, later steps focus on the remaining masked positions. This improves efficiency but can also lock in incorrect tokens. Recent work therefore studies remasking and refinement mechanisms that allow generated tokens to be reconsidered. \citet{wang2025remasking} formulate remasking as inference-time scaling for masked diffusion models. \citet{tian2025finishfirstperfectlater} propose a two-stage fill-then-refine
procedure, while \citet{huang2026dont} train the model to self-reflectively remask incorrect tokens using predicted confidence scores. \citet{zhai2026core} target context-brittle tokens and argue that confidence alone is a poor proxy for correctness. Another line of work improves DLM sampling efficiency by reducing denoising steps, accelerating token updates, or designing better schedules. Fast-dLLM and dInfer accelerate decoding through confidence-based token selection and thresholding \citep{wu2026fastdllm, wu2026fastdllmv2, ma2025dinfer}. Recent information-gain-based sampling methods further suggest that token selection should account for the downstream effect of an update on future denoising uncertainty; we discuss this and other contemporary work in \cref{app:contemporary_sampling}.
\looseness=-1

Our work shares some of the motivation of these remasking and efficient sampling methods, but differs in both signal and allocation target. Rather than applying remasking uniformly or learning a new schedule, we use a trajectory-level frequency signal and apply frequency-aware remasking only to challenging questions.

%% file: sections/setting.tex
\section{Problem Setting}
\label{sec:setting}

We frame test-time scaling as an allocation problem over both compute and token update strategies. To design such an adaptive algorithm, we first formalize test-time scaling for DLMs in \cref{subsec:tts-setting}. We then examine the principle shared by certainty-aware token update strategies and identify a failure mode that motivates allocating additional computation to hard questions in \cref{subsec:false_confidence}.
\looseness=-1

\subsection{Test-Time Scaling for Diffusion Language Models}
\label{subsec:tts-setting}
Let $\mathcal{V}$ denote the token vocabulary and let \texttt{[M]} be a special mask token. A diffusion language model generates a response $y\in\mathcal{Y}$ for a question $x\in\mathcal{X}$ through iterative denoising. Starting from a partially or fully masked sequence, the model repeatedly predicts token distributions for masked positions conditioned on the current sequence state. At denoising step $t$, let $\mathbf{x}^{(t)}\in(\mathcal{V}\cup\{\texttt{[M]}\})^L$ denote the current response state, and let $\mathcal{M}^{(t)}=\{i:x_i^{(t)}=\texttt{[M]}\}$ be the set of masked positions. The model outputs
\[
    p_\theta(x_i \mid \mathbf{x}^{(t)}, x),
    \qquad i\in\mathcal{M}^{(t)},
\]
for all currently masked positions. A decoding strategy $\pi$ then determines which token positions are committed and which positions remain masked for further processing. We write the resulting response as
\[
    y \sim p_\theta^\pi(\cdot \mid x; T),
\]
where $T$ is the number of denoising steps and $\pi$ specifies the token update rule.

Different token update rules induce different tradeoffs between inference cost and generation quality. Confidence-thresholding strategies \citep{wu2026fastdllm, ma2025dinfer} commit tokens once their confidence exceeds a predefined threshold. Intuitively, at each denoising step, the decoder fixes tokens whose confidence is sufficiently high and leaves the remaining tokens available for later processing. We denote this rule abstractly as
\[
    \text{commit token } i
    \quad \Longleftarrow \quad
    \mathrm{conf}_i^{(t)} \geq \delta,
\]
where $\mathrm{conf}_i^{(t)}$ is the model's confidence for position $i$ at step $t$, and $\delta$ is the commitment threshold. Since committed tokens do not need to be repeatedly reconsidered, these strategies can substantially reduce inference cost while maintaining strong average accuracy. In contrast, uncertainty-based remasking strategies, such as low-confidence or high-entropy remasking for \textsc{LLaDA} and \textsc{Dream}, keep uncertain positions active for further processing. These strategies can improve robustness on some hard questions, but they are often significantly slower than confidence-thresholding.

This creates a test-time scaling problem for diffusion decoding. Given a set of questions $\mathcal{S}=\{x_1,\ldots,x_n\}$ and an inference-time budget $B$, we must decide how much compute to spend on each question and which token update strategy to use. Let $c_i$ denote the compute allocated to question $x_i$, measured by inference-time cost such as denoising steps or equivalent forward-pass cost. Let $\pi_i\in\Pi$ denote the strategy selected for question $x_i$, where $\Pi$ is the set of available token update strategies. The goal is to maximize task performance subject to a total compute constraint:
\begin{align}
    \label{eq:dllm_allocation}
    \max_{\{c_i,\pi_i\}_{i=1}^n}
    \quad
    & \frac{1}{n}\sum_{i=1}^{n}
    \mathrm{Perf}(x_i; c_i,\pi_i) \nonumber \\ 
    \text{s.t.}
    \quad
    & \sum_{i=1}^{n} c_i \leq B,
    \qquad c_i\geq 0,\quad \pi_i\in\Pi .
\end{align}
Here, $\mathrm{Perf}(x_i;c_i,\pi_i)$ denotes the evaluation metric, such as coverage or accuracy, obtained by applying strategy $\pi_i$ to question $x_i$ with compute $c_i$ \citep{damani2025learning, manvi2024adaptiveinferencetimecomputellms}. Since the best decoding behavior is not known in advance, the problem is not to choose a single globally best strategy, but to adaptively decide when to \textbf{spend additional compute} and \textbf{when to switch decoding behavior}.
\looseness=-1

\subsection{Certainty-Aware Token Updates}
\label{subsec:false_confidence}

Within this test-time scaling framework, we identify a limitation of widely used certainty-aware token update strategies. A common design principle across diffusion language models and related inference frameworks is to use the model's current predictive distribution to decide which positions should be fixed and which should remain masked for further processing. For each masked position $i \in \mathcal{M}^{(t)}$, the model predicts
\[
    \hat{x}_i^{(t)}
    =
    \argmax_{v \in \mathcal{V}}
    p_\theta(x_i = v \mid \mathbf{x}^{(t)}, x),
\]
and assigns the prediction a confidence score
\[
    \mathrm{conf}_i^{(t)}
    =
    p_\theta(x_i = \hat{x}_i^{(t)} \mid \mathbf{x}^{(t)}, x).
\]
A token update strategy then ranks positions by certainty or applies a threshold to decide which positions are committed and which remain masked.

This formulation captures several existing decoding rules. In LLaDA, low-confidence remasking selects the most confident positions to unmask at each step \citep{nie2025large}. Given a step-dependent number of tokens $k^{(t)}$, the selected positions are
\[
    \mathcal{T}^{(t)}
    =
    \underset{i \in \mathcal{M}^{(t)}}{\mathrm{top}\text{-}k^{(t)}} \; \mathrm{conf}_i^{(t)},
\]
where $\mathcal{T}^{(t)}$ is fixed in the sequence and $\mathcal{M}^{(t)} \setminus \mathcal{T}^{(t)}$ remains masked for later processing. DREAM-style implementations often use entropy-based remasking \citep{ben-hamu2025accelerated,gong2026diffucoder}. Here, token uncertainty is measured by the entropy
\[
    h_i^{(t)}
    =
    - \sum_{v \in \mathcal{V}}
    p_\theta(x_i = v \mid \mathbf{x}^{(t)}, x)
    \log p_\theta(x_i = v \mid \mathbf{x}^{(t)}, x),
\]
where a higher $h_i^{(t)}$ indicates greater uncertainty, and such positions are kept masked for further processing.

These methods therefore rely on instantaneous confidence or entropy as proxies for token certainty. However, local certainty need not imply global correctness: on hard reasoning problems, a token may appear confident because it is locally compatible with an incorrect partial trajectory. We refer to this failure mode as \emph{false confidence} and analyze it in \cref{subsec:fc-analysis}.

%% file: sections/methods.tex
\section{Methods}
\label{sec:methods}

We now present our two-stage adaptive sampling method and the frequency-aware remasking rule used in its second stage. We begin with an analysis of \emph{false confidence} in \cref{subsec:fc-analysis}.
We then define the trajectory-level frequency signal in \cref{subsec:freq}, introduce the SAFE commitment rule built on it in \cref{subsec:safe}, and describe how it is integrated into test-time scaling in \cref{subsec:adaptive_freq}.

\subsection{False Confidence Analysis}
\label{subsec:fc-analysis}
To analyze false confidence, we examine whether failures on hard questions are random or structured. We use \texttt{LLaDA-8B-Instruct} on GSM8K with low-confidence remasking for this analysis. GSM8K is well suited to this analysis because its answers are numerical, which makes final-answer extraction and grouping reliable. We use a diagnostic hard subset of GSM8K identified from failures of LLaDA with confidence-thresholding remasking within 100 samples. For each question in the diagnostic subset, we generate 32 fresh samples and group the extracted final answers after post-processing.
\begin{table}[!ht]
\centering
\caption{
Repeated convergence to wrong answers on unsolved GSM8K questions.
}
\label{tab:wrong_answer_convergence}
\begin{tabular}{lc}
\toprule
Criterion & Fraction of questions \\
\midrule
Dominant wrong answer appears in at least $24/32$ samples & $45.16\%$ \\
Same wrong answer appears in all $32/32$ samples & $14.52\%$ \\
\bottomrule
\end{tabular}
\end{table}

This behavior is consistent with \emph{false confidence}: the model can settle into a locally coherent but globally incorrect reasoning path, causing repeated samples to confidently produce the same wrong final answer. \cref{app:transfer_score} provides a complementary diagnostic that further supports this interpretation. Such failures are difficult to address by simply drawing more samples from the same certainty-aware decoder.

\subsection{Frequency-Aware Remasking}
\label{subsec:freq}

We attribute the failure above to the commitment order: a certainty-aware sampler commits the positions it is most sure about first, and these early commitments steer every sample down the same path. Frequency-aware remasking reverses this order. It tracks how often each position's proposal changes across denoising steps and commits the frequently changing positions first, so that different samples make different choices at the positions where the model is still undecided and branch into different reasoning paths, while the positions the model has already settled on are filled in afterwards. The sampler first follows a certainty-aware warm-up rule during a per-block warm-up phase to collect this statistic, then applies Support-Aware Frequency Selection (SAFE) to choose the high-frequency positions to commit. 
\looseness=-1

For each generated position $i$, we record token proposals while that position remains masked. Let $N_i^{(t)}$ denote the number of proposals observed up to denoising step $t$, including the current
proposal, and let $\hat{x}_{i,j}$ denote the $j$-th observed proposal. We initialize $\hat{x}_{i,0}=[M]$ and define the change frequency as
\[
    f_i^{(t)}
    =
    \frac{
        \sum_{j=1}^{N_i^{(t)}}
        \mathbf{1}\!\left[
            \hat{x}_{i,j} \neq \hat{x}_{i,j-1}
        \right]
    }{
        \max\!\left(N_i^{(t)},1\right)
    }.
\]
This statistic includes the initial comparison against the mask token. A high frequency indicates that a position's proposal changes often, while a low frequency indicates that it remains stable.

Frequency statistics are maintained independently for each generated response and accumulated across its denoising trajectory. They are not reset at block boundaries. Masked positions in future blocks also accumulate observations, although they are not yet eligible for commitment. Once a position is committed, its frequency statistics stop updating.

\paragraph{Warm-up phase.}
Let $S$ denote the prescribed number of denoising steps per block.
We set the number of warm-up steps to
\[
    w
    =
    \min\!\left(
        S-1,\,
        \max\!\left(
            1,\operatorname{round}(\rho S)
        \right)
    \right).
\]
During the first $w$ steps of each block, the sampler follows the warm-up rule while continuing to collect frequency statistics. For example, with 256 total denoising steps distributed evenly across eight blocks, $S=32$, so $\rho=0.125$ gives four warm-up steps per block. When $S=1$, we have $w=0$. The warm-up phase restarts at each block boundary, while the accumulated frequency history is retained. We use $\rho=0.125$ unless otherwise specified.

After warm-up, the sampler must decide which masked positions to commit using the accumulated frequencies $f_i^{(t)}$. A natural choice is to commit every position whose frequency exceeds a fixed threshold. However, the scale of $f_i^{(t)}$ varies across blocks, steps, and tasks, so a fixed threshold requires tuning and may force a selection even when the frequencies show no meaningful separation. We therefore introduce SAFE in \cref{subsec:safe}, which chooses the high-frequency group adaptively from the observed frequency levels.

\subsection{Support-Aware Frequency Selection (SAFE)}
\label{subsec:safe}

First, SAFE looks for a clear separation in the support of the frequency values: the high-frequency group is defined by the largest gap between adjacent raw frequency levels, subject to a size constraint. Second, it commits this group only when it is consistent with the group found at the preceding step in the same block. Whenever either check fails, SAFE falls back to the warm-up rule, so it only overrides that rule when the frequency signal is both well separated and stable.  \looseness=-1

\paragraph{Gap-based candidate selection.}
SAFE is applied after the current proposals have been incorporated into the cumulative frequency statistics. Let $\mathcal{A}_t$ denote the currently masked positions eligible for commitment in the active block, and let their distinct raw frequency values be
\[
    u_1^{(t)} < \cdots < u_m^{(t)}.
\]
When $m\geq3$, SAFE forms a high-frequency candidate for each adjacent
split:
\[
    H_j^{(t)}
    =
    \left\{
        i\in\mathcal{A}_t
        \mid f_i^{(t)}\geq u_{j+1}^{(t)}
    \right\}.
\]
Only splits with a nonempty tail containing at most half of the active positions are admissible:
\[
    \mathcal{J}_t
    =
    \left\{
        j:1\leq j<m,\;
        0<|H_j^{(t)}|\leq|\mathcal{A}_t|/2
    \right\}.
\]
If $\mathcal{J}_t$ is nonempty, SAFE selects
\[
    j_t^\star
    \in
    \operatorname*{arg\,max}_{j\in\mathcal{J}_t}
    \left(u_{j+1}^{(t)}-u_j^{(t)}\right),
    \qquad
    H_t=H_{j_t^\star}^{(t)}.
\]
Ties are resolved in favor of the larger split index, which produces the smaller high-frequency tail. If $m<3$ or $\mathcal{J}_t$ is empty, no valid candidate is formed, and SAFE uses the warm-up rule at that step.

\paragraph{Candidate consistency.}
Let $H_{t-1}$ be the valid candidate from the preceding eligible decision in the same block, and let
\[
    I_t=\mathcal{A}_t\cap\mathcal{A}_{t-1}
\]
contain positions active at both decisions. SAFE accepts $H_t$ only when a preceding candidate exists and
\[
    \left|
        (H_t\cap I_t)
        \mathbin{\triangle}
        (H_{t-1}\cap I_t)
    \right|
    \leq1,
\]
where $\triangle$ denotes symmetric difference. The test therefore allows at most one membership change among positions active at both decisions.

When the consistency test succeeds, SAFE commits all positions in $H_t$. Otherwise, it uses the commitment set of the warm-up rule for that step. The first valid candidate after warm-up establishes the comparison state and therefore uses the warm-up rule. A valid candidate is stored even when it fails the consistency test; if no valid candidate can be formed, the stored candidate is cleared. Candidate state is reset at each block boundary, while frequency statistics remain cumulative across the response.

If the warm-up rule selects no position in an unfinished block, it commits one position according to its own ranking to ensure progress. On the final scheduled step, all remaining positions in the block are committed. The complete per-step procedure is summarized in \cref{alg:safe_step} in \cref{app:safe}.

\begin{algorithm}[t]
\caption{Two-Stage Test-Time Scaling for DLM Sampling}
\label{alg:freq_methods}
\begin{algorithmic}[1]

\renewcommand{\algorithmicrequire}{\textbf{Input:}}
    \renewcommand{\algorithmicensure}{\textbf{Output:}}

\REQUIRE Question set $\cS$, total generation rounds $K$, base-stage rounds
$K_{\mathrm{base}}$, correctness oracle $O:\cX\times\cY\rightarrow\{0,1\}$,
base sampler $\pi_{\mathrm{base}}$, and frequency-aware sampler $\pi_{\mathrm{freq}}$.

\STATE For each question $x \in \cS$, maintain a response set $g(x)$,
a selected response $\check y(x)$, and its correctness value $\check o(x)$.
\STATE Initialize $g(x)\gets\emptyset$, $\check o(x)\gets 0$, and
$\check y(x)\gets\emptyset$ for all $x\in\cS$.
\STATE Initialize the active set $\cA \gets \cS$.

\STATE \textbf{Stage 1: Base DLM sampling.}
\FOR{$r=1$ to $K_{\mathrm{base}}$}
    \IF{$\abs{\cA}=0$}
        \STATE \textbf{break}
    \ENDIF
    \FOR{$x \in \cA$}
        \STATE Generate one response $y \sim \pi_{\mathrm{base}}(\cdot \mid x)$.
        \STATE Update $g(x) \gets g(x) \cup \{y\}$.
        \hfill \algcommentlight{Fast stage: use the base sampler to solve easy questions.}

        \IF{$O(x,y)=1$}
            \STATE Update $\check y(x)\gets y$ and $\check o(x)\gets 1$.
            \STATE Update $\cA \gets \cA \setminus \{x\}$.
            \hfill \algcommentlight{Eliminate once a correct response is found.}
        \ENDIF
    \ENDFOR
\ENDFOR

\STATE \textbf{Stage 2: Frequency-aware DLM sampling.}
\FOR{$r=K_{\mathrm{base}}+1$ to $K$}
    \IF{$\abs{\cA}=0$}
        \STATE \textbf{break}
    \ENDIF
    \FOR{$x \in \cA$}
        \STATE Generate one response $y \sim \pi_{\mathrm{freq}}(\cdot \mid x)$.
        \STATE Update $g(x) \gets g(x) \cup \{y\}$.
        \hfill \algcommentlight{Frequency-aware stage: target unresolved questions.}

        \IF{$O(x,y)=1$}
            \STATE Update $\check y(x)\gets y$ and $\check o(x)\gets 1$.
            \STATE Update $\cA \gets \cA \setminus \{x\}$.
        \ENDIF
    \ENDFOR
\ENDFOR

\ENSURE For each $x\in\cS$, output response set $g(x)$ and selected response $\check y(x)$.
\algcommentlight{A question is solved once $O$ accepts a response; otherwise it remains unsolved.}
\end{algorithmic}
\end{algorithm}

\subsection{Adaptive Test-Time Scaling with Frequency-Aware Remasking}
\label{subsec:adaptive_freq}

We now incorporate frequency-aware remasking into a two-stage test-time scaling procedure, shown in \cref{alg:freq_methods}. Let $\mathcal{S}=\{x_1,\ldots,x_n\}$ denote the set of evaluation questions. For each question $x\in\mathcal{S}$, we maintain a response set $g(x)$ and a selected response $\check{y}(x)$. A correctness oracle $O:\mathcal{X}\times\mathcal{Y}\rightarrow\{0,1\}$ determines whether a generated response correctly solves the question.

The first stage performs adaptive allocation \citep{manvi2024adaptiveinferencetimecomputellms,zuo2026strategic} using a base DLM sampler $\pi_{\mathrm{base}}$, instantiated as confidence-thresholding in our experiments. At each round, the algorithm samples one new response for every active question. If the oracle finds a correct response for a question, that question is removed from the active set. This stage exploits the efficiency of the base sampler and avoids spending additional compute on questions that are already solved.

The second stage applies the frequency-aware sampler $\pi_{\mathrm{freq}}$ only to the remaining active questions. These questions were not solved in the base stage, so the base sampler's commitment order is unlikely to reach a different reasoning path. The frequency-aware sampler $\pi_{\mathrm{freq}}$ denotes a DLM sampler whose warm-up rule is an uncertainty-aware remasking rule (low-confidence for \textsc{LLaDA}, entropy-based for \textsc{Dream}) and which then applies SAFE (\cref{subsec:safe}) in the remaining denoising steps of each block. Thus, frequency-aware remasking is reserved for unresolved questions, rather than being applied uniformly to all inputs. By default, the algorithm performs one inference per question per round, for a total of $K$ rounds.
\looseness=-1

%% file: sections/experiments.tex
\section{Experiments}
\label{sec:experiments}

We describe the experimental setup in \cref{subsec:setup}. We then present test-time scaling results in \cref{subsec:s_results} and a challenging-question analysis in \cref{subsec:u_results}. Ablation studies are provided in \cref{subsec:ablation}, and
additional experimental details and results are deferred to the Appendix.

\subsection{Experimental Setup}
\label{subsec:setup}

\paragraph{Models, baselines, and metrics.}
We evaluate our approach on two representative diffusion language models:
\texttt{LLaDA-8B-Instruct} and \texttt{Dream-7B-Instruct}. For brevity, we refer to them as \textsc{LLaDA} and \textsc{Dream}, respectively. These models allow us to study how different token update and remasking strategies affect iterative decoding under test-time scaling. We report two metrics: coverage@k and accuracy. Coverage@k measures whether at least one of the generated responses for a question is correct after $k$ rounds, as determined by a ground-truth checker. We report coverage rather than pass@$k$ \citep{Chen2021EvaluatingLL} because our adaptive algorithm may allocate different numbers of inference calls to different questions, which violates the fixed-$n$ sampling assumption used by the unbiased pass@$k$ estimator. For mathematical reasoning, we report oracle-measured accuracy: the fraction of questions for which the correctness oracle accepts a response within the allocated budget. We use Gemini-2.5-Flash-Lite to judge each response independently. In \cref{alg:freq_methods}, the first accepted response becomes the selected answer and sampling stops for that question; questions with no accepted response count as unsolved. Thus, when the same oracle is used, this accuracy equals oracle-measured coverage rather than the accuracy of a separate candidate-ranking procedure. Additional oracle-related experiments are presented in \cref{app:oracle}.
\looseness=-1

We evaluate compute budgets of $K\in\{10,20,30,40,50,60\}$ generation rounds per question, with each point in the figures corresponding to one value of $K$. For \textsc{Uniform}, we report results at the same budgets for comparison. For \textsc{Ours (Freq)} and \textsc{Ours (Cert)}, we set $K_{\mathrm{base}}=0.7K$ for $K\in\{10,20,30\}$ and $K_{\mathrm{base}}=0.6K$ for $K\in\{40,50,60\}$, reserving a larger fraction of compute for the second stage at higher budgets. 
Results are averaged over three random runs, and shaded regions in plots denote $\pm 0.5$ standard deviations. Unless otherwise specified, we set the generation length to 256, the number of diffusion steps to 256, and the block length to 32. We compare against standard \textsc{Uniform} scaling and two adaptive baselines: \textsc{Ours (Cert)}, a two-stage certainty-based variant that applies low-confidence remasking for \textsc{LLaDA} and entropy-based remasking for \textsc{Dream} in Stage 2, thereby isolating the contribution of frequency-aware remasking; and \textsc{Elimination} \citep{zuo2026strategic}, a strong adaptive compute-allocation method.

\begin{figure}[t]
  \centering
  \includegraphics[width=.45\linewidth]{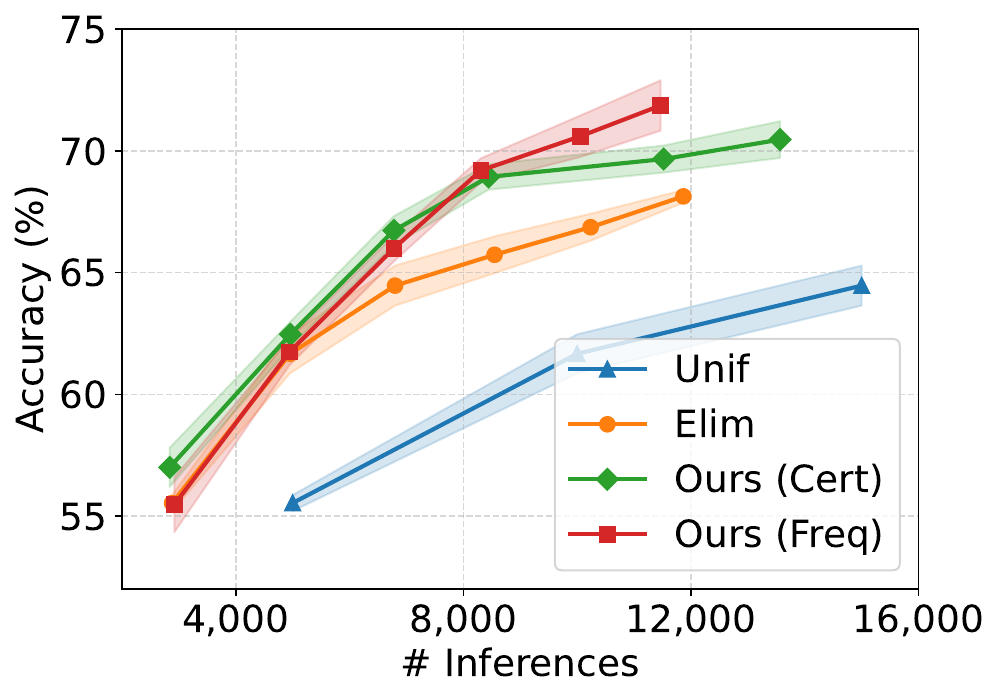}\hfill
  \includegraphics[width=.45\linewidth]{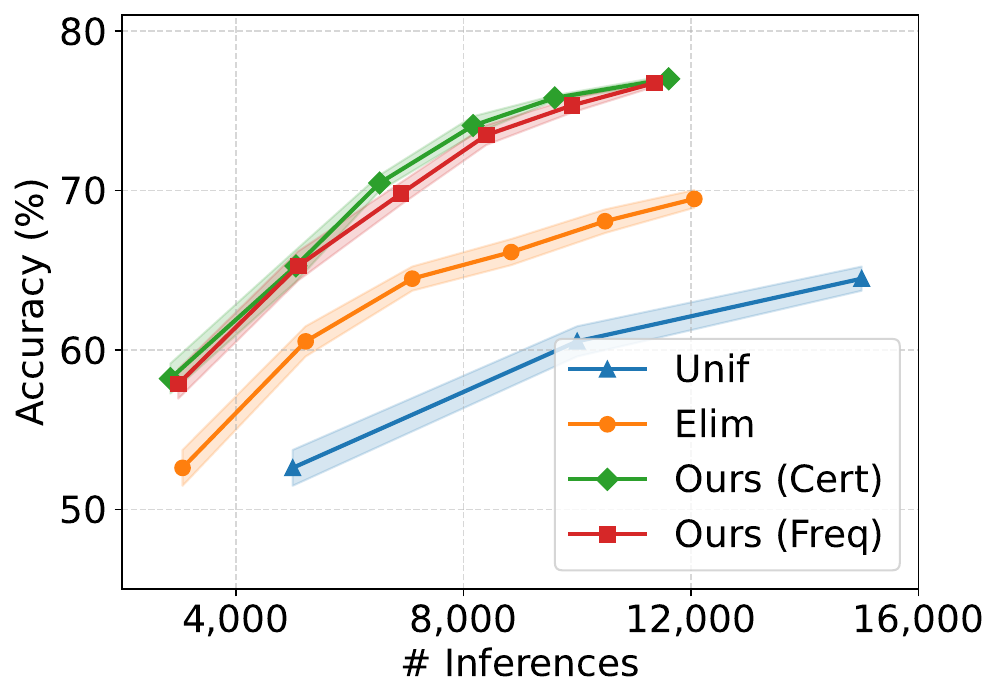}
  \caption{Scaling results on MATH-500.
  \emph{Left:} \textsc{LLaDA}.
  \emph{Right:} \textsc{Dream}.
   Given \cref{alg:freq_methods}, different questions can receive different numbers of inferences under identical $K$.  }
  \label{fig:math500_scale}
  \vspace{-10pt}
\end{figure}

\paragraph{Datasets.}
For the main test-time scaling experiments, we use MATH-500 \citep{lightman2024lets}, which contains 500 competition-level mathematics problems, and HumanEval \citep{Chen2021EvaluatingLL}, which contains 164 programming problems verified by execution tests. For the challenging-question analysis, we additionally use the GSM8K test set \citep{cobbe2021trainingverifierssolvemath}, which contains 1,319 grade-school math word problems, from which we derive a hard subset.

\subsection{Scaling Results}
\label{subsec:s_results}

\paragraph{MATH-500 results.}
We report MATH-500 results in \cref{fig:math500_scale}.
Each point corresponds to a compute budget $K \in \{10,20,30,40,50,60\}$. In this setting, we use an LLM-as-a-judge as the correctness oracle $O$. Across both DLM backbones, \textsc{Ours (Freq)} outperforms the \textsc{Elimination} baseline. For \textsc{LLaDA}, we use a warm-up ratio of $\rho=0.5$. \textsc{Ours (Freq)} performs similarly to \textsc{Elimination} at smaller compute budgets, $K\in\{10,20\}$, but shows increasing gains as the budget grows. For $K\in\{30,40,50,60\}$, frequency-aware remasking provides a consistent advantage, with the largest improvement reaching an absolute accuracy gain of $3.73\%$ at $K=50$. For \textsc{Dream}, we use a warm-up ratio of $\rho=0.125$. Here, the improvement appears earlier and is more pronounced. This is consistent with the observation that, for \textsc{Dream}, entropy-based remasking and confidence-thresholding can differ substantially in performance \citep{li2026breakingfactorizationbarrierdiffusion}. In this case, \textsc{Ours (Freq)} achieves its largest gain at $K=40$, improving accuracy by an absolute $7.33\%$ over \textsc{Elimination}. Compared with \textsc{Ours (Cert)}, which shares the two-stage allocation but keeps the certainty-aware rule in Stage~2, \textsc{Ours (Freq)} is indistinguishable at small budgets and pulls ahead on \textsc{LLaDA} from $K=40$ onward while using fewer inferences. For \textsc{Dream}, the two variants remain within one standard deviation of each other at every budget, so the gain over \textsc{Elimination} on this benchmark comes mainly from the switch of the decoder.
\looseness=-1

\begin{figure}[t]
  \centering
  \includegraphics[width=.45\linewidth]{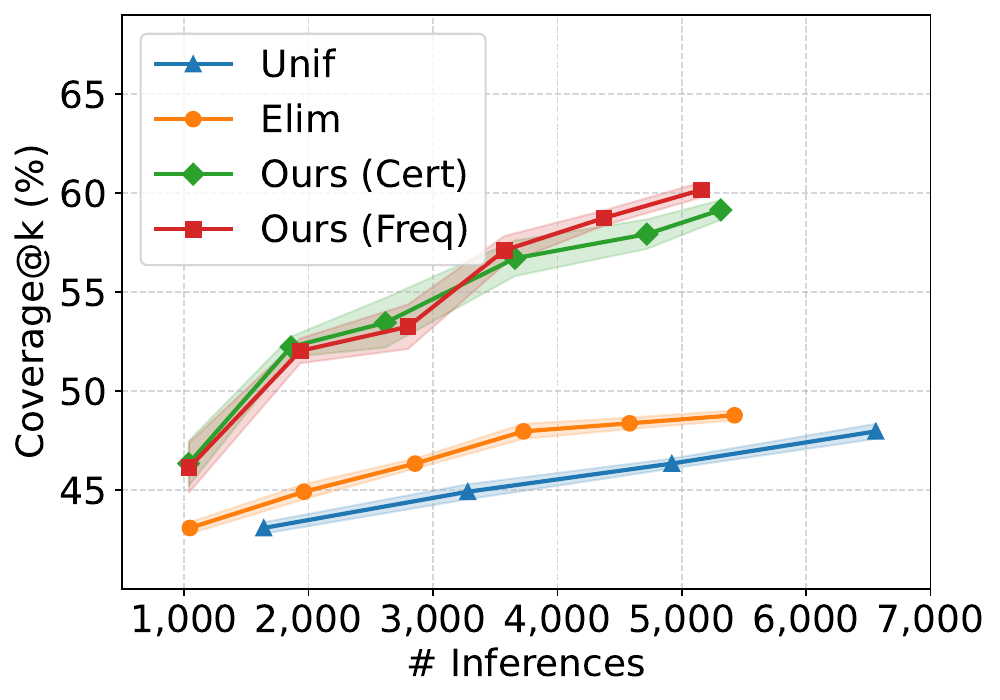}\hfill
  \includegraphics[width=.45\linewidth]{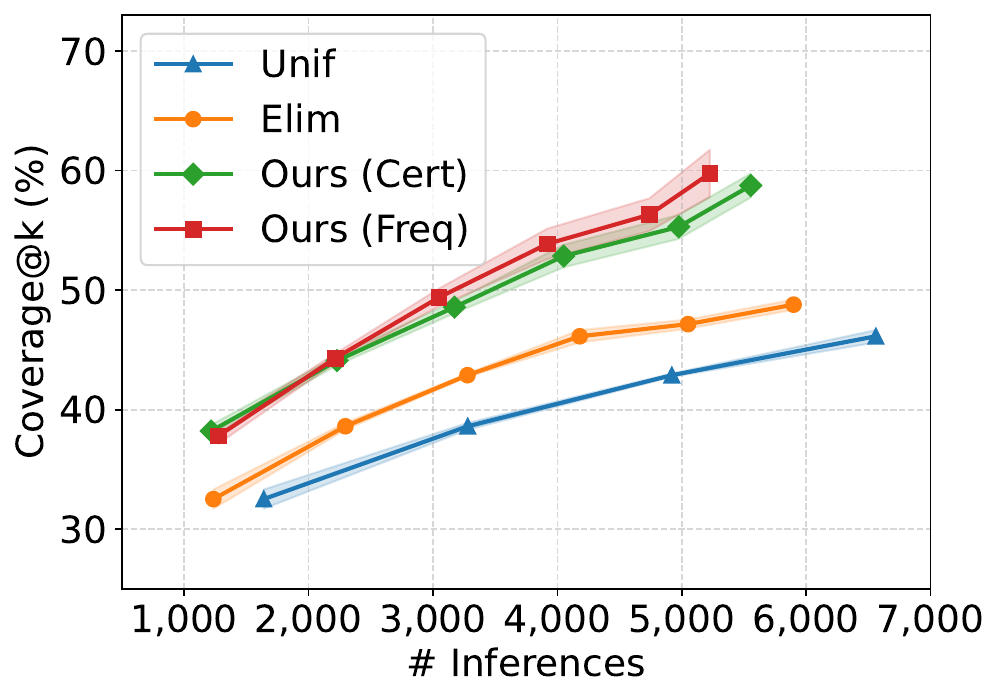}
  \caption{Scaling results on HumanEval.
  \emph{Left:} \textsc{LLaDA}.
  \emph{Right:} \textsc{Dream}.
  Given \cref{alg:freq_methods}, different questions can receive different numbers of inferences under identical $K$.}
  \label{fig:humaneval_scale}
  \vspace{-10pt}
\end{figure}

\paragraph{HumanEval results.}
We report HumanEval results in \cref{fig:humaneval_scale}. Since code generation is directly verifiable, we use the official code execution tests as the correctness oracle $O$ \citep{Chen2021EvaluatingLL}. \textsc{Ours (Freq)} outperforms \textsc{Elimination} on both DLM backbones. For \textsc{LLaDA}, we use a warm-up ratio of $\rho=0.5$. The gains are most pronounced at larger compute budgets: at $K=60$, \textsc{Ours (Freq)} achieves an absolute coverage gain of $11.39\%$ over the baseline. For \textsc{Dream}, we use a warm-up ratio of $\rho=0.125$. In this setting, \textsc{Ours (Freq)} achieves its largest improvement at $K=60$, with an absolute coverage gain of $10.98\%$ over \textsc{Elimination}. On HumanEval, \textsc{Ours (Freq)} also improves over \textsc{Ours (Cert)} at the larger budgets on both backbones, by about one coverage point at $K=60$, while using fewer inferences.

\subsection{Results on Challenging Questions}
\label{subsec:u_results}

We further analyze whether frequency-aware remasking can solve challenging questions that remain unresolved under certainty-based decoding. For each dataset and model, we construct a challenging subset by generating 100 samples per question across three random seeds using the corresponding certainty-based strategy and selecting questions that remain
unsolved. We then evaluate whether frequency-aware remasking can recover solutions on this subset with only 16 samples. For GSM8K, we use a generation length of 128, 128 diffusion steps, and a block length of 32. For each row, the frequency-aware variant retains the same certainty-based strategy during warm-up. The comparison therefore isolates the effect of adding frequency-aware remasking, which differs from our scaling setting. Results are shown in \cref{tab:challenging_all}. Additional experimental details are provided in
\cref{app:setting}.

\begin{table}[!ht]
\centering
\caption{
Representative results on challenging questions. The frequency-aware variant uses the same base strategy as its warm-up strategy and is evaluated with 16 samples. Results are reported as coverage.
}
\label{tab:challenging_all}
\begin{tabular}{llccc}
\toprule
Dataset & Model & Base strategy & Base @100 & + Freq. @16 \\
\midrule
MATH-500 & \textsc{Dream} & Entropy & 0.0\% & 3.75\% \\
HumanEval & \textsc{LLaDA} & Confidence-threshold & 0.0\% & 10.39\% \\
GSM8K & \textsc{LLaDA} & Low-confidence & 0.0\% & 18.75\% \\
\bottomrule
\end{tabular}
\end{table}

Table~\ref{tab:challenging_all} shows that frequency-aware remasking solves challenging questions on all three benchmarks that the aligned certainty-aware strategy leaves unsolved even after 100 samples, with the largest gain for \textsc{LLaDA} on GSM8K.
\textbf{Random} remasking solves almost none of these questions, except for HumanEval with \textsc{LLaDA}, where it reaches 2.60\% coverage at 16 samples, so random exploration alone does not explain the gains. A comparison with temperature sampling and other second-stage alternatives on the GSM8K diagnostic subset, together with answer-diversity statistics, is reported in \cref{tab:answer_diversity} and discussed in \cref{app:e-results}.

\subsection{Ablation Studies}
\label{subsec:ablation}
\begin{figure}[t]
  \centering
  \includegraphics[width=.33\linewidth]{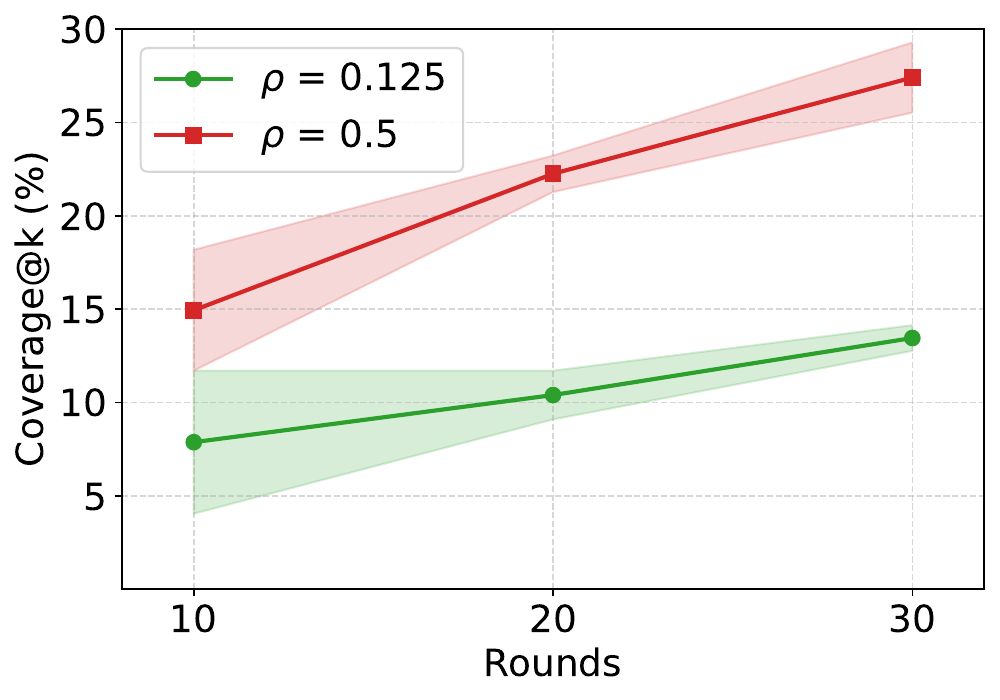}\hfill
  \includegraphics[width=.33\linewidth]{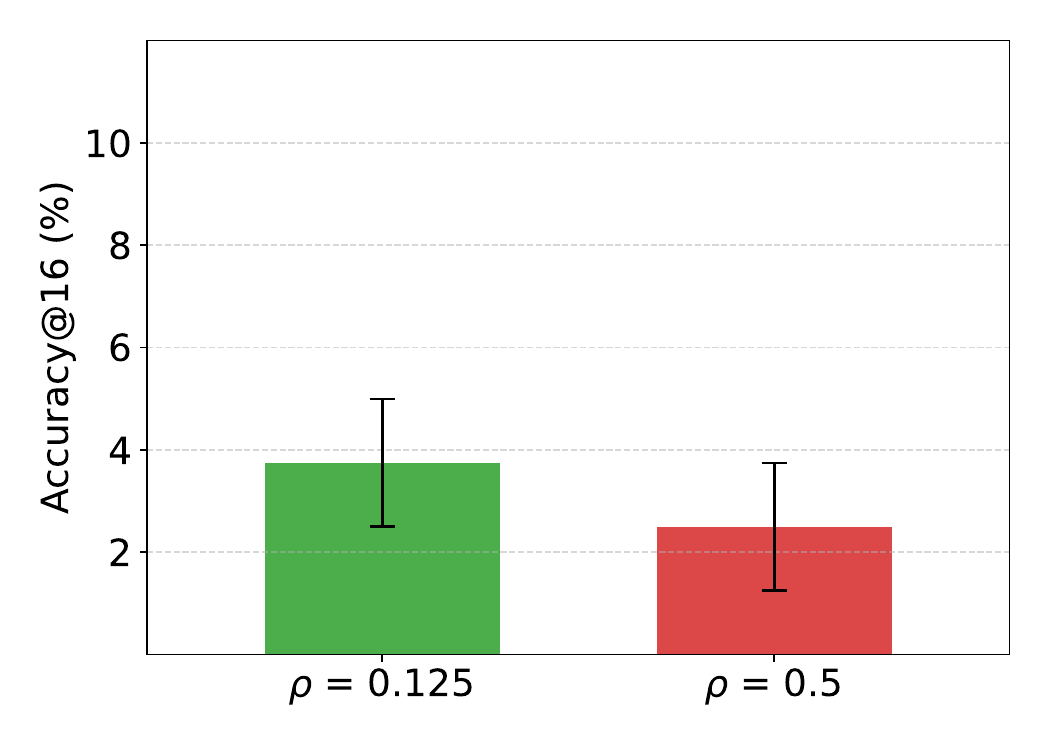}\hfill
  \includegraphics[width=.33\linewidth]{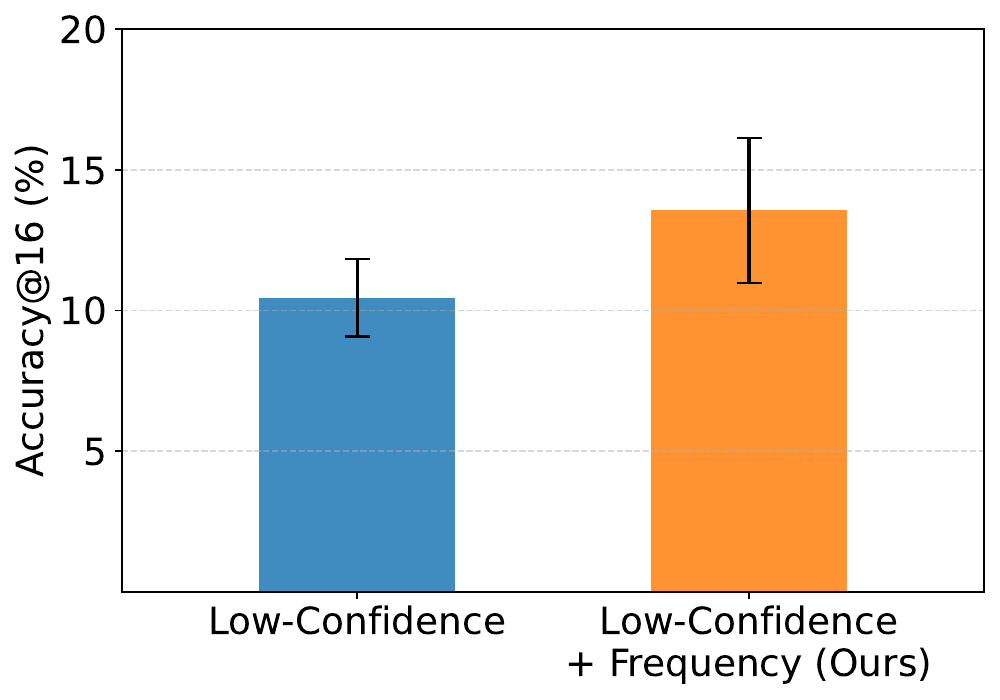}\hfill
  \caption{Ablation studies.
  \emph{Left:} Warm-up ratio $\rho$, case 1.
  \emph{Middle:} Warm-up ratio $\rho$, case 2 (@16).
  \emph{Right:} Frequency-aware remasking on a strong base strategy (@16).}
  \label{fig:ablation}
  \vspace{-10pt}
\end{figure}

\paragraph{Warm-up ratio $\rho$.}
The warm-up ratio $\rho$ controls the fraction of denoising steps that follow the base remasking rule before switching to frequency-aware remasking: a small $\rho$ leaves most steps to the frequency signal, while a large $\rho$ preserves a strong base decoder for longer. We study this tradeoff in two cases. First, \textsc{LLaDA} on the hard HumanEval subset with confidence-thresholding as the base strategy, where the base and frequency-aware rules behave differently and a larger $\rho$ may help preserve the base strategy. Second, \textsc{Dream} on the hard MATH-500 subset with entropy-based remasking, where the two samplers differ only by whether frequency-aware remasking is applied, which isolates the frequency signal. Results in \cref{fig:ablation} show that $\rho$ provides a flexible way to balance base-strategy behavior with frequency-aware remasking.
\looseness=-1

\paragraph{Frequency-aware remasking on a strong base strategy.}
We further test whether frequency-aware remasking can complement an already strong base strategy. We focus on hard MATH-500 questions identified by running \textsc{LLaDA} with confidence-thresholding, and compare standard low-confidence remasking with the same strategy augmented by frequency-aware remasking. This setting tests whether the frequency signal can be added without disrupting the base sampler. Results in \cref{fig:ablation} show that frequency-aware remasking preserves the strength of low-confidence remasking while further improving performance on challenging questions.
\looseness=-1

%% file: sections/discussion.tex
\section{Discussion}
\label{sec:discussion}
We recast DLM test-time scaling as an adaptive allocation problem over two
axes---how many responses to sample and which decoding strategy to apply---departing from the standard view of scaling only sample count under a fixed decoder. Our two-stage algorithm instantiates this view: a fast base stage resolves most questions cheaply, and frequency-aware remasking is applied only to those that remain unresolved. This requires a second-stage decoder that fails differently from the base decoder on the questions the base decoder misses. We identify one source of such failures, \emph{false confidence}, and address it with frequency-aware remasking, a trajectory-level signal based on how often token predictions change across denoising steps. Committing the frequently changing positions first makes different samples branch at the positions where the model is still undecided instead of collapsing onto its preferred reasoning path, which is why the second stage recovers questions that further samples from the base decoder do not. Adapting both where compute is spent and how decoding is performed improves performance on difficult instances while
preserving efficiency. The framework is training-free but introduces hyperparameters such as the warm-up ratio $\rho$. In addition, our signal tracks only changes in token identity, so it is most effective when the difficulty stems specifically from false confidence.
The two-stage formulation is agnostic to the second-stage decoder, leaving richer signals and more adaptive commitment rules to future work.

%% file: sections/appendix.tex
\section{Extended Related Work}
\subsection{Additional Contemporary Work on DLM Sampling and Remasking}
\label{app:contemporary_sampling}

We also discuss several recent works that examine the limitations of local-certainty-based decoding in diffusion language models. \citet{fang2026locallyconfidentgloballystuck} identify a quality--exploration dilemma: low-confidence remasking improves single-sample quality by prioritizing confident tokens, but can suppress exploration and limit multi-sample gains. \citet{yang2026improvingsamplingmaskeddiffusion} argue that greedy local-certainty heuristics ignore how current token choices affect future masked positions, and propose an information-gain sampler that balances immediate uncertainty with downstream uncertainty reduction. \citet{olausson2026tale} study diversity in DLM sampling and introduce tempered variants of confidence-based remasking heuristics to improve pass@$k$ behavior while retaining efficient sampling. \citet{yu2026exploringmorereasoningbetter} instead improve pass@$k$ and answer diversity of DLMs at training time through risk-sensitive GRPO.
\looseness=-1

These works are complementary to ours. They primarily design alternative token-level sampling rules to improve exploration, diversity, or future uncertainty reduction. In contrast, our work studies adaptive allocation across questions and decoding strategies: a fast base sampler is first used to resolve easy questions, and frequency-aware remasking is applied only to unresolved questions. At the token level, our method also uses a different signal---the observed frequency of token prediction changes across denoising steps---rather than only instantaneous confidence, entropy, or estimated information gain. Test-time improvement without additional supervision has also been studied beyond sampling, through self-training at test time \citep{zhu2026testtime} and active evaluation under a labeling budget \citep{liu2026activetestinglargelanguage}.

\section{Additional false confidence analysis}
\label{app:transfer_score}

To support our claim of false confidence, we conduct a diagnostic experiment that tests whether locally confident and stable tokens are always reliable candidates for commitment. Our hypothesis is that, on hard reasoning problems, a token can receive high confidence under the current predictive distribution while still belonging to an incorrect reasoning trajectory.

For each active token position $i$, let $\mathrm{conf}_i$ denote the model confidence and let $f_i$ denote the normalized change frequency of the predicted token across denoising steps. We consider the following confidence-frequency transfer score:
\[
    s_i^{\mathrm{trans}}
    =
    (1-\beta)\,\mathrm{conf}_i - \beta f_i,
\]
where $\beta\in[0,1]$ controls the tradeoff between confidence and frequency. This score prioritizes tokens that are both high-confidence and low-frequency, i.e., tokens whose
predictions are locally confident and stable across the denoising trajectory. If local confidence and trajectory stability were reliable indicators of correctness, then using this score for token commitment should preserve or improve decoding quality. We evaluate this setting using \textsc{LLaDA} on a hard subset of MATH-500 containing 150 questions, with confidence-thresholding as the base strategy and $\beta=0.7$.

\begin{table}[!ht]
\centering
\caption{Comparison between stable-token commitment and our approach.}
\label{tab:stable_vs_ours}
\begin{tabular}{lc}
\toprule
Method & Solved \\
\midrule
Stable-token commitment & 0.00\% \\
Frequency-aware remasking (ours) & 9.33\% \\
\bottomrule
\end{tabular}
\end{table}

As shown in \cref{tab:stable_vs_ours}, the stable-token commitment strategy fails to solve any question in this hard subset, while our frequency-aware remasking strategy solves 9.33\% of the questions. This result suggests that stable high-confidence tokens are not always safe to commit: they may instead reflect locally coherent but globally incorrect reasoning paths. This supports our false-confidence claim and motivates our design choice of committing frequently changing tokens first during the frequency-aware phase, so that samples diverge at the undecided positions rather than collapsing onto the model's preferred reasoning path.

\paragraph{Scope.}
This analysis also clarifies the scope of our method. Frequency-aware remasking is designed to address cases where the model has a viable path to the correct solution but becomes trapped by misleading local certainty during decoding. In such cases, committing the undecided positions first lets different samples branch away from the model's preferred reasoning path. Thus, our method should be understood as a mechanism for recovering from false-confidence failures, rather than a solution for questions that are intrinsically unsolvable for the underlying model.

\section{SAFE Algorithm}
\label{app:safe}
\label{app:safe_algorithm}

\cref{alg:safe_step} gives the per-step commitment procedure of SAFE described in \cref{subsec:safe}. It takes the frequency statistics after the current proposals have been incorporated, together with the warm-up commitment set $B$ and the candidate state from the preceding eligible step in the same block, and returns the positions to commit and the updated candidate state. The candidate state is reset to $(\bot,\emptyset)$ at each block boundary.

\begin{algorithm}[t]
\caption{SAFE Commitment at One Denoising Step}
\label{alg:safe_step}
\begin{algorithmic}[1]
\renewcommand{\algorithmicrequire}{\textbf{Input:}}
\renewcommand{\algorithmicensure}{\textbf{Output:}}

\REQUIRE Active positions $A$; updated frequencies $f$;
warm-up commitment set $B\subseteq A$; previous candidate
$H^{-}$ and active set $A^{-}$; warm-up and terminal-step flags.

\IF{$A=\emptyset$}
    \STATE \textbf{return} $(\emptyset,\bot,\emptyset)$
\ENDIF

\IF{terminal step}
    \STATE \textbf{return} $(A,\bot,\emptyset)$
\ENDIF

\IF{warm-up step}
    \STATE \textbf{return} $(B,\bot,\emptyset)$
\ENDIF

\STATE $U\gets B$ and $H\gets\bot$
\STATE Sort distinct values $\{f_i:i\in A\}$ as
       $u_1<\cdots<u_m$

\IF{$m\geq3$}
    \STATE $H_j\gets\{i\in A:f_i\geq u_{j+1}\}$ for
           $j=1,\ldots,m-1$
    \STATE $J\gets
           \{j:0<|H_j|\leq|A|/2\}$
    \IF{$J\neq\emptyset$}
        \STATE $j^\star\gets
        \arg\max_{j\in J}(u_{j+1}-u_j)$
       \STATE Break ties using the larger $j$
        \STATE $H\gets H_{j^\star}$
    \ENDIF
\ENDIF

\IF{$H\neq\bot$ and $H^{-}\neq\bot$}
    \STATE $I\gets A\cap A^{-}$
    \IF{$|(H\cap I)\mathbin{\triangle}(H^{-}\cap I)|\leq1$}
        \STATE $U\gets H$
    \ENDIF
\ENDIF

\ENSURE Commitment set $U$ and next candidate state $(H,A)$
\end{algorithmic}
\end{algorithm}

\section{Additional Experimental Settings}
\label{app:setting}
\subsection{Hyperparameters and Setup}

We use \texttt{GSAI-ML/LLaDA-8B-Instruct} and \texttt{Dream-org/Dream-v0-Instruct-7B} as the model backbones. Unless otherwise specified, we use a generation length of 256, a prescribed budget of 256 diffusion steps, and a block length of 32. For GSM8K, we use a generation length of 128 and 128 diffusion steps, with the same block length. We use two GSM8K hard subsets, constructed using different base remasking strategies: \cref{tab:wrong_answer_convergence,tab:answer_diversity} use one subset, while the GSM8K results in \cref{tab:challenging_all} use the other. We set the sampling temperature to 0.3 in all experiments except HumanEval with \textsc{Dream}, for which we use 0.4, as discussed in \cref{app:dis}.

We run experiments using existing DLM inference frameworks. Specifically, we use dInfer \citep{ma2025dinfer} for the
\textsc{LLaDA} MATH-500 scaling experiments and Fast-dLLM \citep{wu2026fastdllm} for all remaining experiments. Most experiments are conducted on NVIDIA RTX 6000 Ada Generation GPUs with 48 GB of GPU memory each. Due to GPU memory constraints, we use a batch size of 1 throughout. All evaluations are zero-shot, using either minimal task-specific instructions, such as requesting a boxed final answer, or the question alone.

For frequency-aware remasking, we use SAFE with raw cumulative change frequencies. SAFE does not apply min--max normalization or a fixed frequency threshold. We use $\rho=0.5$ for the \textsc{LLaDA} scaling experiments on MATH-500 and HumanEval, and $\rho=0.125$ for the corresponding \textsc{Dream} experiments. The challenging-question experiments use $\rho=0.125$ for both backbones; the warm-up ablations vary $\rho$ as shown in \cref{fig:ablation}. The warm-up length is computed from the prescribed per-block denoising-step budget using the rounding and clipping rule in \cref{subsec:freq}.

Frequency statistics are collected online and maintained independently for each generated response. For every masked generated position, we accumulate the number of observed proposals and the number of changes relative to the preceding proposal. The previous proposal is initialized to the mask token, so the initial mask-to-proposal comparison is included. The cumulative change frequency is the ratio of the accumulated change count to the accumulated observation count. Frequency statistics are initialized at the start of each response and retained across block boundaries. Masked positions in future blocks also accumulate observations, although they are not yet eligible for commitment. Once a position is committed, its statistics stop updating.

At the beginning of each block, the sampler follows the warm-up rule while continuing to update the frequency statistics. After warm-up, SAFE is evaluated at every nonterminal step. It selects a high-frequency candidate from the largest admissible gap between the observed raw frequency levels and commits the candidate only when it is consistent with the preceding candidate in the same block. Otherwise, it follows the warm-up rule. Candidate history is reset at each block boundary, while frequency statistics remain cumulative across the response. If the warm-up rule selects no position, one position is committed according to its ranking to ensure progress. On the final scheduled step, all remaining positions in the block are committed. Full SAFE details are provided in \cref{subsec:safe}, and the per-step procedure is given in \cref{alg:safe_step}.

\paragraph{Prompt usage.}
\begin{tcolorbox}[
    colback=gray!5,
    colframe=gray!50,
    title={MATH-500 Prompt},
    fonttitle=\bfseries,
    breakable
]
\begin{verbatim}
Put the final answer in \boxed{ }.
Problem:
{question}

Solution:
\end{verbatim}
\end{tcolorbox}

For Dream's HumanEval setting, we use the official \citep{wu2026fastdllm} setting, which is adapted from the official \citep{ye2025dream} GitHub repo.

\begin{tcolorbox}[
    colback=gray!5,
    colframe=gray!50,
    title={LLaDA's HumanEval Prompt},
    fonttitle=\bfseries,
    breakable
]
\ttfamily
Complete the following Python function. Return only Python code.\\[1mm]
\{HumanEval prompt\}
\end{tcolorbox}

\begin{tcolorbox}[
    colback=gray!5,
    colframe=gray!50,
    title={GSM8K Prompt},
    fonttitle=\bfseries,
    breakable
]
\ttfamily
\{question\}
\end{tcolorbox}

\section{Observations from Our Experiments}
\label{app:dis}

Since DLMs are relatively new, there are still few well-established best practices for decoding and evaluation. We therefore summarize several observations from our experimental process. All observations are based on the settings described in \cref{app:setting}.

First, we find that \textsc{LLaDA} is relatively robust to temperature and confidence threshold choices compared with \textsc{Dream}. For confidence thresholds in the range $0.90$--$0.95$, \textsc{LLaDA} shows stable single-sample performance, and we obtain results comparable to those reported in \citet{nie2025large} on both HumanEval and
MATH-500. In contrast, \textsc{Dream} is much more sensitive to the confidence threshold: changing the threshold from $0.90$ to $0.95$ can shift single-sample performance from roughly $\sim 20\%$ to $\sim 30\%$ in our setting.

We also observe that \textsc{Dream} is sensitive to temperature. Taking HumanEval as an example, we use the official evaluation script released in the Fast-dLLM GitHub repository, which is adapted from the official Dream model repository. Under this setting, temperature has an unexpected effect. When the temperature is set to $0$, the model produces
normal outputs. However, when the temperature is set in the range $0.1$--$0.3$, the generation often collapses or fails to produce usable outputs, leaving coverage below $10\%$ after post-processing. Interestingly, this issue diminishes when the temperature is increased to the range $0.4$--$1.0$: although performance is lower, the outputs return to a reasonable range. This behavior may be related to how \textsc{Dream} handles temperature in its source code. Based on this observation, we use temperature $0.4$ for this HumanEval setting only.

\section{Additional Experimental Results}
\label{app:e-results}

\paragraph{Oracle.}
\label{app:oracle}
\cref{tab:llada_math500} compares the ground-truth oracle with the LLM judge on MATH-500 with \textsc{LLaDA}. Agreement between the two oracles exceeds 93\% for \textsc{LLaDA} and 95\% for \textsc{Dream}, producing nearly identical baseline accuracy. Gemini-2.5-Flash-Lite ranks second on mathematics among reward models in \citet{malik2026rewardbench}, so we consider it a close proxy for ground-truth correctness. The LLM judge is slightly more lenient than the ground truth.
\looseness=-1

\begin{table}[!ht]
\centering
\caption{Oracle comparison for \textsc{LLaDA} on MATH-500: accuracy (\%) under ground-truth (GT) and LLM judging. These results are reported separately from the three-run scaling averages in \cref{fig:math500_scale}. The ground-truth and LLM-judged results for \textsc{Elimination} coincide at every budget.}
\label{tab:llada_math500}
\begin{tabular}{lcccccc}
\toprule
Method & @10 & @20 & @30 & @40 & @50 & @60 \\
\midrule
Elim (GT)  & 54.8 & 59.4 & 62.6 & 64.2 & 65.8 & 67.4 \\
Elim (LLM) & 54.8 & 59.4 & 62.6 & 64.2 & 65.8 & 67.4 \\
Ours (GT)  & 54.2 & 59.8 & 64.8 & 68.2 & 69.4 & 71.0 \\
Ours (LLM) & 53.2 & 60.0 & 64.8 & 68.4 & 69.6 & 71.4 \\
\bottomrule
\end{tabular}
\end{table}

\paragraph{Second-stage alternatives on hard questions.}
We compare SAFE with four second-stage alternatives using \textsc{LLaDA} on the GSM8K diagnostic subset from \cref{subsec:fc-analysis} (\cref{tab:answer_diversity}). \emph{Low-confidence} is plain low-confidence remasking with no second-stage change, and serves as the reference for the other rows, all of which use it as their warm-up rule. \emph{Temperature sampling} keeps low-confidence remasking and raises the sampling temperature to $T=1.2$. \emph{Low-frequency-first} is a counterfactual that uses the same warm-up and commits the same number of positions per step as a fixed-threshold frequency rule, but draws them from the lowest-frequency positions instead of the highest. \emph{Confidence dynamics} also matches the per-step commitment count of the frequency rule, but ranks positions by the mean absolute change of their top-1 confidence between consecutive steps rather than by how often their proposal changes. At 16 samples, SAFE achieves the highest coverage. Confidence dynamics is the strongest alternative, which indicates that committing positions whose predictions are still moving helps whether the movement is measured by proposal changes or by confidence changes, while committing the most stable positions first is no better than raising the temperature. The 18.75\% recovery rate in \cref{tab:challenging_all} is measured on the separately derived recovery subset.

\paragraph{Answer diversity.}
Frequency-first commitment changes what the samples do, not only how many succeed. On the same diagnostic subset, we also count the distinct final answers among the 16 samples of each question and the share of samples that give the most common answer (\cref{tab:answer_diversity}). Low-confidence remasking produces few distinct answers and a dominant answer that most samples repeat, consistent with collapse onto the model's preferred reasoning path. Temperature sampling, low-frequency-first commitment, and confidence dynamics each add diversity, while SAFE roughly quadruples the number of distinct answers and spreads the samples across them. Diversity alone is not sufficient: low-frequency-first produces as many distinct answers as confidence dynamics but recovers far fewer questions, so the branching must happen at positions where the model is still undecided.

\begin{table}[H]
\centering
\caption{Second-stage strategies on the GSM8K hard subset with \textsc{LLaDA} at 16 samples. \emph{Distinct answers}: mean number of different final answers per question. \emph{Dominant share}: mean fraction of samples giving the most common answer. \emph{Coverage@16}: fraction of questions with at least one correct answer.}
\label{tab:answer_diversity}
\begin{tabular}{lccc}
\toprule
Method & Distinct answers & Dominant share (\%) & Coverage@16 (\%) \\
\midrule
Low-confidence & 2.4 & 78.8 & 17.74 \\
Temperature sampling ($T=1.2$) & 5.3 & 58.8 & 25.81 \\
Low-frequency-first & 7.7 & 45.5 & 25.81 \\
Confidence dynamics & 7.5 & 47.1 & 37.10 \\
SAFE & \textbf{10.2} & \textbf{31.2} & \textbf{45.16} \\
\bottomrule
\end{tabular}
\end{table}